\documentclass[10pt,journal,compsoc,twoside]{IEEEtran}
\usepackage{times}
\usepackage{amsmath}
\usepackage{amssymb}
\usepackage{algorithm}
\usepackage{algorithmic}
\usepackage{graphicx}
\usepackage{multirow}
\usepackage{subfigure}
\usepackage{color}
\usepackage[english]{babel}
\usepackage{array}

\ifCLASSOPTIONcompsoc
  \usepackage[nocompress]{cite}
\else
  \usepackage{cite}
\fi
\ifCLASSINFOpdf
\else
\fi

\begin{document}

\title{Subdomain Adaptation with Manifolds Discrepancy Alignment}

\author{Pengfei Wei$^1$, Yiping Ke$^2$, Xinghua Qu$^2$, Tze-Yun Leong$^1$\\
National University of Singapore$^1$,
Nanyang Technological University$^2$.}


\IEEEtitleabstractindextext{
\begin{abstract}
Reducing domain divergence is a key step in transfer learning problems.
Existing works focus on the minimization of global domain divergence.
However, two domains may consist of several shared subdomains, and differ from each other in each subdomain.
In this paper, we take the local divergence of subdomains into account in transfer.
Specifically, we propose to use low-dimensional manifold to represent subdomain, and align the local data distribution discrepancy in each manifold across domains.
A Manifold Maximum Mean Discrepancy (M3D) is developed to measure the local distribution discrepancy in each manifold.
We then propose a general framework, called Transfer with Manifolds Discrepancy Alignment (TMDA), to couple the discovery of data manifolds with the minimization of M3D.
We instantiate TMDA in the subspace learning case considering both the linear and nonlinear mappings.
We also instantiate TMDA in the deep learning framework.
Extensive experimental studies demonstrate that TMDA is a promising method for various transfer learning tasks.
\end{abstract}

}

\maketitle

\IEEEdisplaynontitleabstractindextext
%
\IEEEpeerreviewmaketitle

\IEEEraisesectionheading{\section{Introduction}}
\IEEEPARstart{L}{abeled} data scarcity is one of the challenges in conventional machine learning as label acquisition is an expensive and time-consuming process \cite{Bishop2006prml}.
To alleviate such a label scarcity issue, there is a strong motivation to take advantage of the labeled data that have been collected in previous tasks.
Precisely, given a new target domain with none or limited labeled data, it is desired to learn a model that leverages the rich labeled data from a different but related source domain.
However, due to the domain divergence between the source and target domains, the learned model may suffer from a poor generalization capability.

Transfer learning is a learning paradigm that can efficiently bridge the domain divergence across different domains \cite{pan2010survey}.
It has been widely used in many real-world applications, e.g., sentiment analysis \cite{Wei2017Hybrid}, affective computing \cite{li2019multisource}, visual object recognition \cite{wang2019softly}, signal processing \cite{xie2018generalized}, etc.
Various methods including subspace-based methods \cite{gong2012geodesic,sun2016return,Wei2017Hybrid} and deep learning methods \cite{Wei2016DeepNF,Long2017Deep,Cao2018Partial} have been proposed.
These methods share a similar idea, that is, aligning different domains by reducing the domain divergence through appropriate feature mappings.
Domain-invariant feature representations are learned and then used to transfer knowledge across domains.

Although effective in some transfer learning tasks, existing methods mainly focus on the alignment of \emph{global domain divergence}, i.e., the global data distribution discrepancy of two domains.
However, two domains may consist of several shared subdomains \cite{Li2018Domain}.
The local subdomain divergence, i.e., the local data distribution discrepancy of subdomains, may not be well aligned in the transfer process.
This may lead to a degeneration of the discriminative power of the learned domain invariant features, and even worse, negative transfer.
For instance, in many NLP tasks \cite{dai2007co}, a document could contain several hidden topics.
Each topic can be taken as a subdomain.
The same hidden topic may have different word distributions in different domains.
In the transfer process, it is more reasonable to align the data within the same hidden topic across domains individually than to align all the data together.
This is because the latter may easily mix up the discriminative words of different hidden topics in the alignment, and thus degenerate the transfer performance.

To align subdomain divergence, one essential step is to define sudomains.
Class or category of data is one way to do so.
In this case, the alignment of subdomains enforces the data within the same class to be closer, and thus helps with positive transfer.
However, using class to define subdomains may be not applicable or optimal when the target domain does not have any labeled data or only has limited labeled data.
Existing works \cite{long2013transfer,Bachman2014Learning,zhang2015deep} propose to utilize pseudo target labels.
These methods highly rely on the initialization of pseudo target labels, and suffer from the issue of error reinforcement, where mistakenly assigned target labels lead to the distribution alignment of wrong classes.

To avoid the problematic pseudo target labelling, in this paper, we propose to utilize a new way, i.e., low-dimensional manifold, to define subdomains.
According to \cite{Vidal2010Subspace}, data points of many real-world domains are very likely to be sampled from a distribution supported by multiple low-dimensional manifolds embedded in the ambient space.
Thus, a domain can be divided into several local subdomains where each subdomain contains the data lying in one low-dimensional manifold.
For instance, in the above NLP tasks, different low-dimensional manifolds could represent different hidden topics.
Precisely, in sentiment analysis, a domain is a set of reviews of a product that are labeled by sentiment classes (positive, neutral, and negative).
Customers express their altitude to a product from different aspects (latent topics), e.g., the appearance, the price, or the functionality etc.
Each aspect corresponds to one low-dimensional manifold, and it is natural to align samples that fall within the same manifold across domains.
In practice, using low-dimensional manifold to define subdomain is generally applicable to various transfer learning problems as (1) it does not require target labels, and (2) it has intuitive interpretation in various real-world tasks, e.g., text-based tasks \cite{blitzer2007biographies,dai2007co} where each manifold may correspond to one hidden topic, and visual-based tasks \cite{gong2012geodesic,long2013transfer} where each manifold may correspond to one object in images.
\begin{figure}[!t]
\centering
\resizebox{\linewidth}{!}{
{\includegraphics[scale = 0.36]{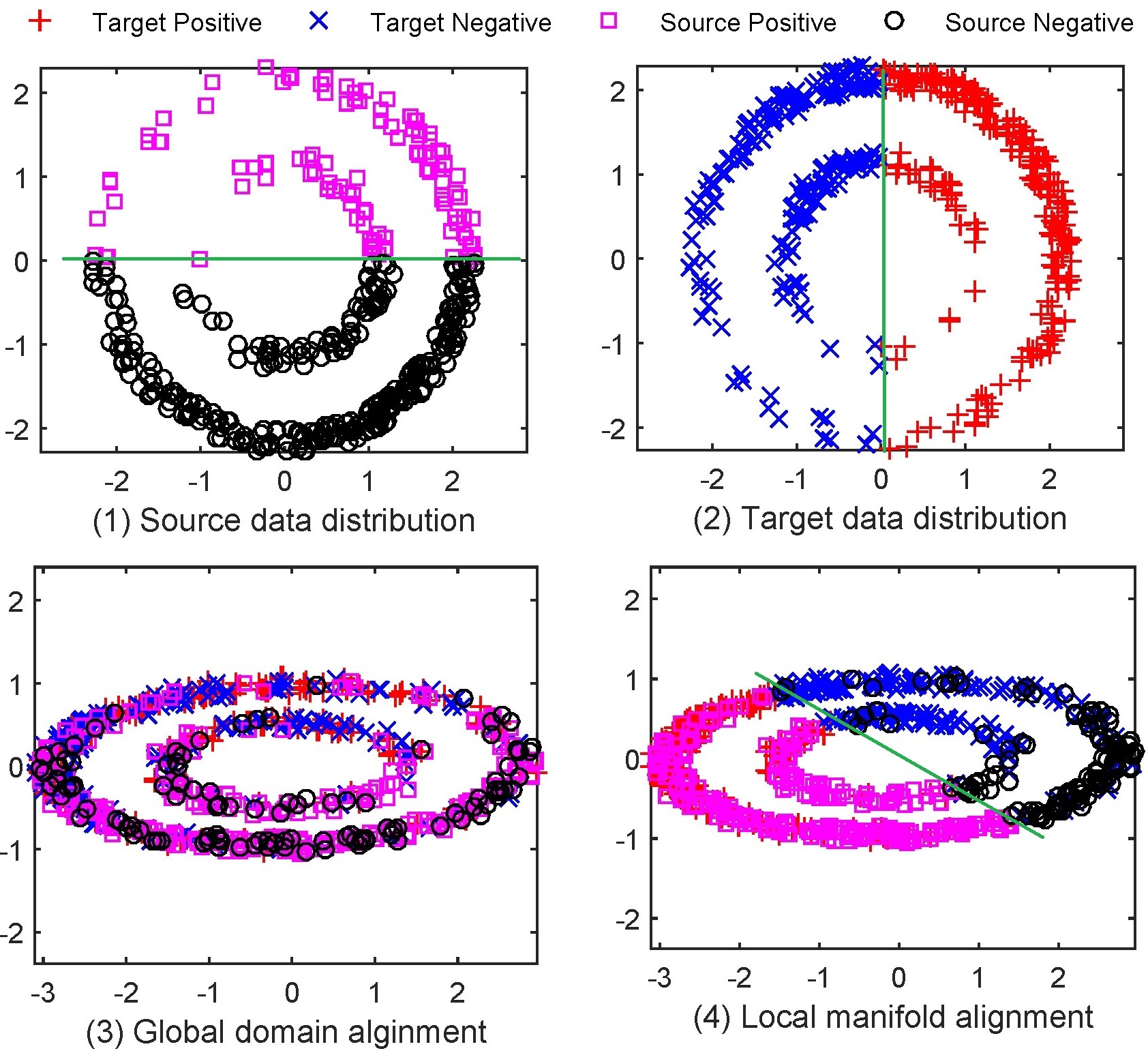}}
}
 \caption{An example. The magenta squares, the black circles, the red `+'s and the blue `x's correspond to the source positive, source negative, target positive and target negative samples. The green line is the discriminative hyperplane.} \label{example}
\end{figure}

Figure \ref{example} gives a visual example of aligning domains in each manifold. The data of two domains lie in the same manifolds, i.e., two annuli, but draw different distributions as shown in figure \ref{example} (1) and (2).
Note that we assume a simple 2-dimensional space and keep the dimensionality unchanged in the domain alginment for the visualization purpose.
If the global domain alignment is done, e.g., by reducing the global distribution discrepancies, the discriminative structure of the two domains are mixed up as shown in figure \ref{example} (3).
However, if we algin the domains in each annulus, the source and target data can be well classified by a shared hyperplane as shown in figure \ref{example} (4).
In this paper, we aim to reduce the local subdomain divergence by minimizing the distribution discrepancy of different domains in each low-dimensional manifold.
To do so, we propose a general framework, called Transfer with Manifolds Discrepancy Alignment (TMDA), that couples the data low-dimensional manifold discovery with the minimization of subdomain divergence in a unified structure.
In order to measure the data distribution discrepancy in each low-dimensional manifold, we first propose a new local metric, called Manifold Maximum Mean Discrepancy (M3D).
We compare our proposed M3D with the conventional Maximum Mean Discrepancy (MMD) on various transfer learning tasks.
The experimental results show that M3D significantly improves the transfer performance compared with MMD.
Using M3D, we instantiate the general TMDA framework in both subspace learning and deep learning cases.
Extensive experimental studies show the effectiveness of our TMDA framework on various transfer tasks.

This work can be taken as an extension of the conference paper \cite{wei2019knowledge} published in CIKM 2019.
Compared with \cite{wei2019knowledge}, this work makes following new contributions.
Firstly, we propose a new idea of aligning subdomains for transfer in this work, and take advantage of multiple manifolds to represent subdomains. This is different from the idea of \cite{wei2019knowledge} that aims to transfer multiple manifolds information, although both of the two works are based on multiple manifolds assumption.
Secondly, we propose a general sudomain discrepancy alignment framework (TMDA), which consists of manifolds discovery term, manifolds discrepancy term and regularization term. TMDA allows to utilize different techniques to achieve these 3 terms.
Thirdly, we develop a new local metric, M3D, which can measure the local subdomain divergence across different domains, and can be used as the manifolds discrepancy term in the TMDA framework.
Fourthly, we instantiate our TMDA in both subspace learning and deep learning scenarios. Compared with \cite{wei2019knowledge}, we add the nonlinear mapping case for subspace learning scenario, and also achieve the instantiation for deep learning scenario.
Finally, we conduct more empirical experiments, on both synthetic and real-world datasets, to verify our TMDA framework.

\section{Related Works}
Subspace learning methods have been shown effective for transfer learning tasks.
MMD-based method is one of the most popular subspace learning methods.
In \cite{pan2008transfer}, MMD is investigated to find a common latent subspace by dimensionality reduction methods.
In the extended studies, Pan et al. \cite{pan2011domain} propose a transfer component analysis (TCA) algorithm to learn the shared features in a reproducing kernel Hilbert space (RKHS).
Following studies combine MMD with other techniques to further boost transfer.
Long et al. \cite{long2013transfer} put forward a joint distribution adaptation method to align both the marginal and conditional distributions simultaneously.
Transfer joint matching (TJM) \cite{long2014transfer} reduces domain divergence by jointly minimizing MMD and reweighting instances.
A recent subspace-based work, MEDA \cite{wang2018visual}, combines both the marginal and conditional MMDs with the geometric structure learning, and demonstrates a superiority to deep methods on some adaptation tasks.
Wei et al. \cite{wei2019knowledge} propose to integrate the global domain alignment with the local manifold neighborhood structure preservation in the transfer procedure.
Except for MMD-based methods, subspace alignment based methods are also widely studied.
Gong et al. \cite{gong2012geodesic} propose to generate intermediate subspaces along the geodesic path between source and target subspaces on the Grassmann manifold.
To alleviate its computational issue, Fernando et al. propose to align the source and target subspaces directly \cite{fernando2013unsupervised}.
Other methods also consider important data properties in transfer, e.g., data locality \cite{shu2014locality}, second-order statistics \cite{sun2016return}, and geometric discriminative structure \cite{zhang2017joint}.

Recently, deep learning methods have attracted increasing attention in transfer learning as deep features can disentangle complex and high-level information underlying data, making the feature representations more discriminative and informative to transfer tasks.
Chen et al. \cite{chen2012marginalized} propose a marginalized stacked denoising autoencoder (mSDA) method that marginalizes out random feature corruptions, and it achieves significant positive transfer performance on the cross domain sentiment analysis.
In \cite{Wei2016DeepNF}, Wei et al. introduce MMD and kernalization to mSDA to further boost the transfer capability.
Long et al. \cite{long2015learning} develop a deep adaptation network (DAN) to learn transferable features.
Multiple kernelized MMD is adopted to reduce domain discrepancies in the last three layers of DAN.
Different from DAN, DTN \cite{zhang2015deep} minimizes marginal distributions in each layer using MMD, and it also minimizes conditional distributions in the last discriminative layer by using pseudo target labels iteratively.
In \cite{Long2017Deep}, a joint adaptation network (JDN) is proposed to minimize the joint distribution of full-layers features through MMD.
Considering the class weight bias across domains, Yan et al. \cite{Yan2017Mind} develop a weighted domain adaptation network (WDAN), which improves the transfer performance of DAN.
Although generally more effective than subspace-based methods, many deep transfer learning methods are actually motivated from subspace-based ones.

\section{Problem Setting}
We are given a source domain ${\mathcal{D}}_s = \{(\mathbf{x}_i^s,y_i^s)\}_{i=1}^{n_s}$ with $n_s$ labeled data, and a target domain ${\mathcal{D}}_t = \{\mathbf{x}_i^t\}_{i=1}^{n_t}$ with $n_t$ unlabeled data.
The feature spaces of the two domains are the same, but the feature distributions are different, specifically, $p_s$ for the source domain and $p_t$ for the target domain.
The source data matrix is denoted as $\mathbf{X}_s \in \mathbb{R}^{d \times n_s}$ and its corresponding label vector is $\mathbf{y}_s \in \mathbb{R}^{1 \times n_s}$.
Similarly, the target data matrix is denoted as $\mathbf{X}_t \in \mathbb{R}^{d \times n_t}$.
Moreover, we denote the joint data matrix as $\mathbf{X} = [\mathbf{X}_s, \mathbf{X}_t] \in \mathbb{R}^{d \times n}$ ($n = n_s + n_t$) with $\mathbf{x}_i \in \mathbb{R}^{d \times 1}$.
The source and target domains consist of $N$ shared subdomains corresponding to $N$ low-dimensional manifolds $\{\mathcal{M}_i \in \mathbb{R}^{d_i}\}_{i=1}^N$ which are embedded in the original feature space $\mathbb{R}^d \ (d_i < d)$.
We further denote the source and target data matrices potentially embedded in the $m$-\emph{th} low-dimensional manifold $\mathcal{M}_m$ as $\mathbf{X}_s^m \in \mathbb{R}^{d \times n_s^m}$ and $\mathbf{X}_t^m \in \mathbb{R}^{d \times n_t^m}$, which are drawn from the distributions $p_s^m$ and $p_t^m$, respectively.

\section{A General Transfer Framework with Manifolds Discrepancy Alignment}
\subsection{A General Framework}
To reduce the subdomain divergence, a key step is to discover the low-dimensional manifolds given the source and target data matrix $\mathbf{X}$.
Manifold discovery, also known as manifold clustering, is a widely studied topic, and a comprehensive survey including several branches of manifold clustering algorithms can be found in \cite{Vidal2010Subspace}.
Herein, we are interested in the spectral clustering-based methods that explore a spectral graph $\mathcal{G}$ over $\mathbf{X}$.
Multiple manifolds can be obtained by applying the ncut clustering algorithm \cite{Shi2000Normalized} on $\mathcal{G}$.
To couple the manifold discovery with the local discrepancy minimization, we propose a general framework, called Transfer with Manifolds Discrepancy Alignment (TMDA), with the objective function:
\begin{equation} \label{Eq1}
\mathcal{O}_{tmda}(\phi(\cdot),\mathcal{G}) = \mathcal{MD}(\phi(\cdot),\mathcal{G}) + \lambda \hat{d}^\prime(\phi(\cdot),\mathcal{G}) + \mathcal{R}(\phi(\cdot)),
\end{equation}
where $\mathcal{MD}$ is the manifold discovery term, $\hat{d}^\prime$ is the manifold discrepancy term, $\mathcal{R}$ is the regularization term and $\phi(\cdot)$ is the feature mapping to be learned.
Specifically, the objective of $\mathcal{MD}$ is to discover the multiple manifolds embedded in the data and simultaneously preserve the manifolds information in the new feature representation learning.
The objective of $\hat{d}^\prime$ is to minimize the local distribution discrepancy in each low-dimensional manifold based on $\mathcal{G}$.
The objective of $\mathcal{R}$ is to regularize $\phi(\cdot)$.
Note that the TMDA objective of Eq. (\ref{Eq1}) is general in the sense that (1) it allows different ways to achieve the three terms, and (2) it is applicable to both subspace learning and deep learning.
Regarding the first point, various methods, e.g., locally linear manifold clustering \cite{goh2007segmenting}, sparse subspace clustering \cite{elhamifar2009sparse}, or low-rank representation \cite{Liu2010Robust}, can be adjusted to construct an affinity matrix $\mathbf{A}$ to represent $\mathcal{G}$.
Distance metrics, e.g., Maximum Mean Discrepancy (MMD) \cite{borgwardt2006integrating}, KL-divergence \cite{rosenberg2001color}, A-distance \cite{ganin2016domain}, can be used as the base metric to be further improved for $\hat{d}^\prime$.
Similarly, $\mathcal{R}$ can be any regularization including $l_1$ norm, $l_2$ norm, $l_{21}$ norm and $l_{rank}$ norm etc.
Regarding the second point, $\phi$ can be a single feature mapping matrix for the subspace learning case, or multiple-layer network weights for the deep learning case.
In the following sections, we instantiate each element of TMDA in both subspace and deep learning cases.

\subsection{Manifold Maximum Mean Discrepancy}
We start with the key element $\hat{d}^\prime$.
Herein, we focus on the metric MMD as it is widely used in both subspace and deep transfer learning works.
It is a nonparametric metric to quantify the discrepancy of two distributions.
It takes the mean embeddings of the two distributions in a Reproducing Kernel Hilbert Space (RKHS) $\mathcal{H}$ as a distance calculation to avoid the density estimation.
Formally, given the source and target data $\mathbf{X}_s$ and $\mathbf{X}_t$, drawn from the distributions $p_s$ and $p_t$ respectively, MMD is defined as:
\begin{equation} \nonumber
d(p_s,p_t) \triangleq ||\mathbf{E}_{p_s}(\phi(\mathbf{x}_i^s)) - \mathbf{E}_{p_t}(\phi(\mathbf{x}_i^t))||_{\mathcal{H}}^2,
\end{equation}
where $\phi(\cdot)$ is the feature mapping function that maps the original data to RKHS.
The empirical MMD is given below:
\begin{equation}\label{Eq2}
\begin{aligned}
\hat{d}(p_s,p_t) &= {\rm{||}}\frac{1}{n_s}\sum\limits_{i = 1}^{n_s} {\phi ({{\mathbf{x}}_i^s}) - } \frac{1}{n_t}\sum\limits_{i = 1}^{n_t} {\phi ({{\mathbf{x}}_i^t})} |{|^2_{\mathcal{H}}} \\
& = \frac{1}{n_s^2} \sum\limits_{i = 1}^{n_s} \sum\limits_{j = 1}^{n_s} k(\mathbf{x}_i^s,\mathbf{x}_j^s) + \frac{1}{n_t^2} \sum\limits_{i = 1}^{n_t} \sum\limits_{j = 1}^{n_t} k(\mathbf{x}_i^t,\mathbf{x}_j^t) \\
& - \frac{2}{n_s n_t} \sum\limits_{i = 1}^{n_s} \sum\limits_{j = 1}^{n_t} k(\mathbf{x}_i^s,\mathbf{x}_j^t),
\end{aligned}
\end{equation}
where $k(\mathbf{x}_i^s,\mathbf{x}_i^t) = \langle \Phi(\mathbf{x}_i^s),\Phi(\mathbf{x}_i^t) \rangle$ and $\langle \cdot,\cdot \rangle$ is the inner product operator.
Considering the local subdomain divergence, we aim to quantify the distribution discrepancy of data within the same low-dimensional manifold across domains.
To do so, we propose a Manifold Maximum Mean Discrepancy (M3D) metric as:
\begin{equation} \nonumber
d^\prime(p_s,p_t) \triangleq \mathbf{E}_m ||\mathbf{E}_{p_s^m}(\phi(\mathbf{x}_i^{s(m)})) - \mathbf{E}_{p_t^m}(\phi(\mathbf{x}_i^{t(m)}))||_{\mathcal{H}}^2,
\end{equation}
where $\mathbf{x}_i^{s(m)}$ and $\mathbf{x}_i^{t(m)}$ are the data lying in the $m$-\emph{th} low-dimensional manifold, and $p_s^m$ and $p_t^m$ are the distributions of $\mathbf{X}_s^m$ and $\mathbf{X}_t^m$, respectively.
Correspondingly, the empirical M3D is defined as:
\begin{equation} \label{Eq3}
\begin{aligned}
\hat{d}^\prime(p_s,p_t) &= \frac{1}{N}\sum\limits_{m = 1}^{N}{\rm{||}}\frac{1}{n_s^m}\sum\limits_{i = 1}^{n_s^m} {\phi ({{\mathbf{x}}_i^{s(m)}}) - } \frac{1}{n_t^m}\sum\limits_{i = 1}^{n_t^m} {\phi ({{\mathbf{x}}_i^{t(m)}})} |{|^2_{\mathcal{H}}} \\
& = \frac{1}{N}\sum\limits_{m = 1}^{N}(\sum\limits_{i = 1}^{n_s^m} \sum\limits_{j = 1}^{n_s^m} \frac{1}{{n_s^m}^2}k(\mathbf{x}_i^{s(m)},\mathbf{x}_j^{s(m)}) \\
& + \frac{1}{{n_t^m}^2} \sum\limits_{i = 1}^{n_t^m} \sum\limits_{j = 1}^{n_t^m} k(\mathbf{x}_i^{t(m)},\mathbf{x}_j^{t(m)})\\ &- \frac{2}{n_s^m n_t^m} \sum\limits_{i = 1}^{n_s^m} \sum\limits_{j = 1}^{n_t^m} k(\mathbf{x}_i^{s(m)},\mathbf{x}_j^{t(m)})).
\end{aligned}
\end{equation}
Compared Eq. (\ref{Eq2}) with Eq. (\ref{Eq3}), it can be observed that MMD only measures the global distribution discrepancy of two domains, while M3D could measure the discrepancy of local distributions, specifically in each low-dimensional manifold.
Note that although the form of M3D is similar to that of Conditional Maximum Mean Discrepancy (CMMD) \cite{long2013transfer}, M3D is significantly different from CMMD as the former uses low-dimensional manifolds to define subdomains while the latter uses class.

\subsection{TMDA Instantiation in Subspace Learning}
In the subspace learning case, the feature mapping can be a transformation matrix mapping the data from the original space to a subspace.
We denote the feature mapping function as $\Phi(\cdot)$, and propose the following objective function:
\begin{equation} \label{Eq4}
\begin{split}
    {\mathop {\min}\limits_{\Phi,\bf{A}}} \ \ & \underbrace{\frac{1}{2}||\mathbf{X}-\mathbf{X}\mathbf{A}||_F^2 + \mu||\mathbf{A}||_{1}
    +\frac{\alpha}{2} ||\Phi(\mathbf{X})-\Phi(\mathbf{X})\mathbf{A}||_F^2}_{Manifolds \ Discovery} \\ &+ \underbrace{\frac{\beta}{2} tr(\Phi(\mathbf{X})({\sum\limits_{m=1}^N\mathbf{M}^m})\Phi(\mathbf{X})^\mathsf{T})}_{Manifolds \  Discrepancy} + \underbrace{\frac{1}{2}||\Phi||^2_{F}}_{Regularization} \\
    s.t.{\rm{\ \ }} &diag(\mathbf{A}) =\mathbf{0}, \  \Phi(\mathbf{X})\Phi(\mathbf{X})^\mathsf{T} = \mathbf{I},
\end{split}
\end{equation}
where
\begin{equation} \nonumber
\begin{split}
\mathbf{M}_{ij}^m = \left\{ {\begin{array}{ll}
 \frac{1}{n_s^m n_s^m}, \ \ \ \ \mathbf{x}_i, \mathbf{x}_j \in \mathbf{X}_s^m, \\
  \frac{1}{n_t^m n_t^m}, \ \ \ \   \mathbf{x}_i, \mathbf{x}_j \in \mathbf{X}_t^m , \\
  -\frac{1}{n_s^m n_t^m}, \ \ \  \mathbf{x}_i \in \mathbf{X}_s^m, \mathbf{x}_j \in \mathbf{X}_t^m , \\
  \ \ \ \ \ \ \ \ \ \ \ \ \ \ \ \ or \ \mathbf{x}_i \in \mathbf{X}_t^m, \mathbf{x}_j \in \mathbf{X}_s^m , \\
   0, \ \ \ \ otherwise .\\
\end{array}} \right.
\end{split}
\end{equation}
Objective (\ref{Eq4}) is an instantiation of (\ref{Eq3}).
The first three terms correspond to $\mathcal{MD}$.
The fourth term corresponds to $\hat{d}^\prime$.
The last term corresponds to $\mathcal{R}$.
The constraint is to exclude arbitrary scaling factors in the feature transformation.
More specifically, the first two terms are based on sparse representation, which assumes that each data point in a union of manifolds can be efficiently reconstructed by a sparse combination of other points that are lying in the same manifold with itself.
According to \cite{elhamifar2009sparse}, there exists a sparse solution $\mathbf{A}$ that captures such a relationship of points in the same manifold.
The third term aims to preserve the manifold information, i.e., $\mathbf{A}$, from the original data to the new feature representations.
The fourth term is the trace form of M3D that is to reduce the local distribution discrepancy in each manifold between two domains.
The last term is using Frobenious regularization to control the complexity of the feature transformation.

Regarding $\Phi(\cdot)$, a straightforward choice, also the most widely used one in the existing works, is using a linear transformation matrix $\mathbf{W}$, i.e.,
\begin{equation} \label{Eq5}
\Phi(\mathbf{X}) = \mathbf{W}^\mathsf{T}\mathbf{X}.
\end{equation}
Herein, we consider a more general solution taking both the linear and nonlinear feature mapping into account, that is, a kernelized solution.
Specifically, we first map the original data to a RKHS $\mathcal{H}$ through a mapping function $\Psi$ where ${k}({\bf{x}}_i,{\bf{x}}_j)=\Psi({\bf{x}}_i)^{\mathsf{T}}\Psi({\bf{x}}_j)$ is the kernel of $\mathcal{H}$, and then learn a linear mapping function in $\mathcal{H}$.
According to \cite{niyogi2004locality}, the linear mapping function in $\mathcal{H}$ can be represented as a linear combination of the data points in $\mathcal{H}$, that is, $ {\mathbf{W}}^{\mathsf{ T}}{\Psi {({\bf{X}})}}$.
In this case, we have:
\begin{equation} \nonumber
\Phi(\mathbf{X}) = \mathbf{W}^\mathsf{T}\Psi(\mathbf{X})\Psi(\mathbf{X})^\mathsf{ T} = \mathbf{W}^\mathsf{T} \mathbf{K}.
\end{equation}
where $\mathbf{W} \in \mathbb{R}^{k \times n}$ with $k$ as the dimensionality of the subspace to be learned, and $\mathbf{K} \in \mathbb{R}^{n \times n}$ is the kernel matrix calculated using the kernel function $k(\cdot,\cdot)$ on the input data $\mathbf{X}$.
Correspondingly, we have the following kernelized objective function:
\begin{equation} \label{Eq6}
\begin{split}
    {\mathop {\min}\limits_{({\bf{W}},\bf{A})}} & \frac{1}{2}||\mathbf{X}-\mathbf{X}\mathbf{A}||_F^2 + \mu||\mathbf{A}||_1  +\frac{\alpha}{2} ||\mathbf{W}^{\mathsf{T}}\mathbf{K}-\mathbf{W}^{\mathsf{T}}\mathbf{K}\mathbf{A}||_F^2 \\
    &+ \frac{\beta}{2} tr(\mathbf{W}^{\mathsf{T}}{\bf{K}}({\sum\limits_{m=1}^N\mathbf{M}^m}){{\bf{K}}^{\mathsf{T}}}\mathbf{W}) + \frac{1}{2}tr(\mathbf{W}^{\mathsf{T}}\mathbf{W}) \\
    s.t.{\rm{\ \ }} &diag(\mathbf{A}) =\mathbf{0}, \  {{\bf{W}}^{\mathsf{T}}}{\bf{K}}{{\bf{K}}^{\mathsf{T}}}{\bf{W}} = \mathbf{I}.
\end{split}
\end{equation}
We can utilize linear kernel and nonlinear kernel for (\ref{Eq6}) to achieve the linear mapping and nonlinear mapping, respectively.
Note that, one can also use Eq. (\ref{Eq5}) for the linear mapping case.
The objective is then changed by simply replacing the kernel matrix $\mathbf{K}$ with the input data matrix $\mathbf{X}$.

To solve the optimization problem (\ref{Eq6}), we propose to use Alternating Direction Method of Multipliers (ADMM) method \cite{boyd2011distributed} to iteratively optimize $\mathbf{A}$ and $\mathbf{W}$.
Specifically, given the variables ($\mathbf{W}_p$, $\mathbf{A}_p$) of the $p$-\emph{th} iteration ($p \in [1,maxIter1]$), we optimize these variables for the $(k+1)$-\emph{th} iteration.
Firstly, we optimize $\mathbf{A}_{p+1}$ by fixing $\mathbf{W}_p$.
This is equivalent to solve:
\begin{equation} \label{Eq7}
\begin{split}
    &{\mathop {\min}\limits_{\bf{A}}} \frac{1}{2}||\mathbf{X}-\mathbf{X}\mathbf{A}||_F^2 + \mu||\mathbf{A}||_1 + \frac{\alpha}{2} ||\mathbf{W}_p^{\mathsf{T}}\mathbf{K}-\mathbf{W}_p^{\mathsf{T}}\mathbf{K}\mathbf{A}||_F^2 \\
    & s.t. \ diag(\mathbf{A}) =\mathbf{0}.
\end{split}
\end{equation}
We use the optimal solution of (\ref{Eq7}) as $\mathbf{A}_{p+1}$. To solve (\ref{Eq7}), we introduce an auxiliary matrix $\mathbf{Z} \in \mathbb{R}^{n \times n}$ into (\ref{Eq7}), and add a penalty term corresponding to the constraint.
We then derive the Lagrangian function by introducing a matrix $\mathbf{\Delta} \in \mathbb{R}^{n \times n}$ of Lagrange multipliers to the constraint.
This gives us:
\begin{equation} \label{Eq8}
\begin{split}
    \mathcal{L}^\prime(\mathbf{A},\mathbf{Z},\mathbf{\Delta}) &=   \frac{1}{2}||\mathbf{X}-\mathbf{X}\mathbf{Z}||_F^2 + \mu||\mathbf{A}||_1 \\&+ \frac{\alpha}{2} ||\mathbf{W}_p^{\mathsf{T}}\mathbf{K}-\mathbf{W}_p^{\mathsf{T}}\mathbf{K}\mathbf{Z}||_F^2
   \\& +  \frac{\rho}{2}||\mathbf{Z}-(\mathbf{A}-Diag(\mathbf{A}))||_F^2 \\&+ tr((\mathbf{A}-Diag(\mathbf{A})-\mathbf{Z})\mathbf{\Delta}).
\end{split}
\end{equation}
To obtain the optimal $\mathbf{A}_{p+1}$, we then iteratively update $\mathbf{A}$, $\mathbf{Z}$, and $\mathbf{\Delta}$.
Denote the $q$-\emph{th} iteration optimization variables as ($\mathbf{A}^{q}$, $\mathbf{Z}^q$) and the $q$-\emph{th} iteration Lagrange multipliers as $\mathbf{\Delta}^q$ ($q \in [1,maxIter2]$), we obtain:

\noindent (1) $\mathbf{Z}^{q+1}$ by minimizing $\mathcal{L}^\prime$ with respect to $\mathbf{Z}$, while ($\mathbf{A}^{q}$,$\mathbf{\Delta}^q$) are fixed. By setting the derivative of $\mathcal{L}^\prime$ with respect to $\mathbf{Z}$ to be zero, we obtain:
\begin{equation} \label{Eq9}
\begin{split}
(\mathbf{X}^\mathsf{T}\mathbf{X} + \rho\mathbf{I} + \alpha\mathbf{K}^\mathsf{T}\mathbf{W}_p\mathbf{W}_p^\mathsf{T}\mathbf{K})\mathbf{Z}^{q+1} &=  \mathbf{X}^\mathsf{T}\mathbf{X} + \rho\mathbf{A}^q + \mathbf{\Delta}^q \\
& + \alpha\mathbf{K}^\mathsf{T}\mathbf{W}_p\mathbf{W}_p^\mathsf{T}\mathbf{K}.
\end{split}
\end{equation}
$\mathbf{Z}^{q+1}$ can be obtained by solving Eq. (\ref{Eq9}).

\noindent (2) $\mathbf{A}^{q+1}$ by minimizing $\mathcal{L}^\prime$ with respect to $\mathbf{A}$, while ($\mathbf{Z}^q$,$\mathbf{\Delta}^q$) are fixed.
The optimization problem with respect to $\mathbf{A}$ is a standard $l_1$-norm optimization problem \cite{beck2009fast} and it has a closed-from solution:
\begin{equation} \label{Eq10}
\begin{split}
\mathbf{A}^{q+1} = \mathbf{J} - Diag(\mathbf{J}),\\
\mathbf{J} \triangleq \mathcal{T}_\mu(\mathbf{Z}^{q+1} - {\mathbf{\Delta}^q}/{\rho}),
\end{split}
\end{equation}
where $\mathcal{T}_\mu(v) = (|v|-\mu)_{+}sgn(v)$.
The operator $(\cdot)_{+}$ returns the larger value between the argument and zero.

\noindent (3) $\mathbf{\Delta}^{q+1}$ with ($\mathbf{A}^{q+1}$,$\mathbf{Z}^{q+1}$) fixed by using:
\begin{equation} \label{Eq11}
\mathbf{\Delta}^{q+1} = \mathbf{\Delta}^{q} + \rho(\mathbf{Z}^{q+1}-\mathbf{A}^{q+1}).
\end{equation}

The iteration stops when it converges or the number of iterations reaches $maxIter2$.
The convergence is achieved when $||\mathbf{A}^{q} - \mathbf{Z}^q||_F^2 \leq \epsilon$ and $||\mathbf{A}^{q} - \mathbf{A}^{q-1}||_F^2 \leq \epsilon$, where $\epsilon$ is a predefined error tolerance.
Then we use the obtained optimal $\mathbf{A}^{opt}$ as $\mathbf{A}_{p+1}$:
\begin{equation} \label{Eq12}
\mathbf{A}_{p+1} = \mathbf{A}^{opt}.
\end{equation}

Next, we optimize $\mathbf{W}_{p+1}$ by fixing $\mathbf{A}_{p+1}$.
This is to solve the following optimization problem:
\begin{equation} \label{Eq13}
\begin{split}
   {\mathop {\min }\limits_{\bf{W}}} {\rm{\  }} &   \alpha ||\mathbf{W}^{\mathsf{T}}\mathbf{K}-\mathbf{W}^{\mathsf{T}}\mathbf{K}\mathbf{A}_{p+1}||_F^2 + tr(\mathbf{W}^{\mathsf{T}}\mathbf{W})\\ & + \beta tr(\mathbf{W}^{\mathsf{T}}{\bf{K}}(\sum\limits_{m=1}^N\mathbf{M}^m){{\bf{K}}^{\mathsf{T}}}\mathbf{W}) \\
    s.t.{\rm{\ \ \ \ }} &{{\bf{W}}^{\mathsf{T}}}{\bf{K}}{{\bf{K}}^{\mathsf{T}}}{\bf{W}} = \mathbf{I}.
\end{split}
\end{equation}
Note that to obtain $\mathbf{M}^m, \ m \in [1,N]$, we conduct the ncut clustering algorithm \cite{Shi2000Normalized} on $\mathbf{A}_{p+1}$ to expose the $N$ manifolds.
Optimization objective (\ref{Eq13}) can be formulated into the following generalized eigenvalue problem:
\begin{equation} \label{Eq14}
( \mathbf{I} + \mathbf{K} (  \sum\limits_{m=1}^N \beta\mathbf{M}^m + \alpha \mathbf{N}_{p+1})\mathbf{K}^\mathsf{T})\mathbf{W} = \mathbf{K} \mathbf{K}^{\mathsf{T}} \mathbf{W} \mathbf{\Psi}_{p+1}.
\end{equation}
where $\mathbf{N}_{p+1} = (\mathbf{I}-\mathbf{A}_{p+1} )(\mathbf{I}-\mathbf{A}_{p+1} )^\mathsf{T}$ and $\mathbf{\Psi}_{p+1}$ are the lagrange multipliers.
The optimal solution $\mathbf{W}_{p+1}$ can be derived by solving the top $s$ smallest eigenvectors from Eq. (\ref{Eq14}).

The iteration of update with respect to $\mathbf{A}$ and $\mathbf{W}$ stops when the convergence is achieved or the number of iterations reaches $maxIter1$.
The convergence is achieved when we have $||\mathbf{A}_{p}- \mathbf{A}_{p-1}||_F^2 \leq \epsilon$ and $||\mathbf{W}_{p} - \mathbf{W}_{p-1}||_F^2 \leq \epsilon$.
Finally, we use the obtained optimal $\mathbf{W}_{opt}$ for transfer.

Regarding the hyper-parameters $\mu, \rho, \alpha$ and $\beta$, we introduce the following configuration instructions.
The hyper-parameter $\mu$ is to weight the sparsity.
As suggested in \cite{kim2007interior}, we set:
\begin{equation} \nonumber
\mu = \mathop{{\min}}\limits_{i} \mathop{{\max}}\limits_{i \neq j} |\mathbf{x}_i^\mathsf{T}\mathbf{x}_j|.
\end{equation}
For $\rho$ that weights the penalty term in ADMM, we use the default setting $\rho=1$ as suggested in \cite{boyd2011distributed}.
For $\alpha$ and $\beta$ that balance the effects of manifolds discovery and local discrepancy minimization, we conduct sensitivity analysis, and give an empirical configuration.


\subsection{TMDA Instantiation in Deep Learning}
So far, we have instantiated the general TMDA in subspace learning.
In this section, we discuss how to achieve TMDA in deep learning scenarios.
We start with a formal formulation of deep transfer network which is widely used in existing works \cite{long2015learning,zhang2015deep,Long2017Deep,Yan2017Mind}:
\begin{equation} \label{eq17}
{\mathop {\min }\limits_{f}} \ \frac{1}{n_s} \sum\limits_{i=1}^{n_s} \mathcal{L}(f(\mathbf{x}_i^s),y_i^s) + \lambda \hat{d}(p_s,p_t),
\end{equation}
where $f$ indicates the learnable weights of the deep network, $\mathcal{L}$ is the discriminative loss on the source labeled data, and $\hat{d}$ is the global transfer loss.
To apply TMDA to the deep network, we may consider to add $\mathcal{MD}$ term to (\ref{eq17}) and replace the global transfer loss $\hat{d}$ with the local transfer loss $\hat{d}^\prime$.
Consequently, we have:
\begin{equation} \label{eq18}
\begin{aligned}
{\mathop {\min }\limits_{f,\mathbf{A}}} &\ \frac{1}{n_s} \sum\limits_{i=1}^{n_s} \mathcal{L}(f(\mathbf{x}_i^s),y_i^s) + \lambda_1 \mathcal{MD}(f,\mathbf{A}) + \lambda_2 \hat{d}^\prime(f,\mathbf{A}),\\
\end{aligned}
\end{equation}
An instantiation of (\ref{eq18}) is using sparse representation to $\mathcal{MD}$ as the subspace learning case does.
Specifically, we define:
\begin{equation} \nonumber
\mathcal{MD}(f,\mathbf{A}) =\frac{1}{2}||\mathbf{X}-\mathbf{X}\mathbf{A}||_F^2 + \mu||\mathbf{A}||_{1}
    +\frac{\alpha}{2} ||f(\mathbf{X})-f(\mathbf{X})\mathbf{A}||_F^2.
\end{equation}
Regarding $\hat{d}^\prime$, we can either use M3D or use the adversarial loss.
For M3D, we simply apply the proposed Eq. (\ref{Eq3}) to $\hat{d}^\prime$.
For the adversarial loss, we can train a domain discriminator using the cross entropy loss in each manifold.
After the formulation, the optimization is to iteratively solve $f$ and $\mathbf{A}$ in every iteration.
Given $f$, we update $\mathbf{A}$ using similar steps in the subspace learning case.
Given $\mathbf{A}$, we may use the standard backpropagation to update the network weights $f$.
It is worth noting that our TMDA framework can be built upon the existing deep transfer learning methods with the similar objective function in (\ref{eq17}).
In this work, we instantiate our TMDA for deep learning case based on the benchmark method DAN \cite{long2015learning}.

\section{Experimental Studies}
In this section, we first demonstrate the superiority of M3D to the conventional MMD on several synthetic datasets. Then, we evaluate our proposed TMDA on 4 real-world transfer learning benchmark datasets.

\subsection{Synthetic Experiments}
We first compare the proposed M3D with the conventional MMD on a set of synthetic transfer learning tasks.
To do this, we test on two domains which consist of multiple subdomains, i.e., multiple low dimensional manifolds.
Following \cite{Liu2010Robust}, we construct data from 5 independent manifolds $\{\mathcal{M}_i\}_{i=1}^5$.
The bases $\{\mathbf{U}_i\}_{i=1}^5$ of $\{\mathcal{M}_i\}_{i=1}^5$ are computed by $\mathbf{U}_{i+1} = \mathbf{T}\mathbf{U}_i, 1 \leq i \leq 4$, where $\mathbf{T}$ is a random rotation and $\mathbf{U}_1$ is a random
orthogonal matrix of dimension $\mathbb{R}^{100 \times 10}$.
Thus, the ambient space has a dimension of 100, and each manifold has a dimension of 10.
Both the source and target domains share $\{\mathcal{M}_i\}_{i=1}^5$.
However, we sample the source and target data using different distributions.
Specifically, We sample 40 data points from each manifold by $\mathbf{X}_i = \mathbf{U}_i\mathbf{Q}_i,  1 \leq i \leq 5$, where $\mathbf{Q}_i$ is a sampling matrix, for each domain.
We define $\mathbf{Q}_i$ as a $10 \times 40$ i.i.d. $\mathcal{N}(0.05,0.1)$ for the source domain, but a $10 \times 40$ i.i.d. $\mathcal{N}(-0.05,0.1)$ for the target domain.
Some data points are then randomly corrupted by a random gaussian noise.
We then obtain 200 source data points and 200 target data points, embedded in the same manifolds but with different distributions.
For both domains, we assign a label to each data point based on the manifold where the data is from.
Thus, the two domains share the same label space $\{1,2,3,4,5\}$.
Note that labels are assumed to be unknown for the target domain in transfer.
Following these steps, we generate 10 datasets corresponding to 10 transfer tasks, denoted as $\mathcal{D}_1,...,\mathcal{D}_{10}$.

We test M3D and MMD on $\mathcal{D}_1,...,\mathcal{D}_{10}$.
Specifically, we use TCA \cite{pan2011domain}, which only minimizes MMD in the transfer process, as the baseline.
We compare TCA using MMD ($T_{mmd}$) with TCA using M3D ($T_{3md}$) on $\mathcal{D}_1,...,\mathcal{D}_{10}$.
Note that we obtain M3D by using \cite{elhamifar2009sparse} to firstly cluster the data into $N$ manifolds and then calculate the trace form of M3D according to Eq. (\ref{Eq3}).
Other experimental configurations are set to be the same for $T_{mmd}$ and $T_{3md}$.
We also consider both non-kernalized and kernelized cases.
For the non-kernelized case, we use Eq. (\ref{Eq5}) as the linear mapping.
For the kernelized case, we follow \cite{pan2011domain} and use `linear' kernel for the linear mapping case and `rbf' kernel for the nonlinear mapping case.
As suggested by \cite{gong2012geodesic}, we utilize the Nearest Neighbor (NN) classifier as the base classifier since it does not require tuning cross-validation parameters.
We use root mean squared error (RMSE) as the evaluation metric.
\begin{table}[t]
\centering
\caption{Comparison results on $\mathbf{D}_i, i =1,...,10$ with $N=5$}
\label{ARCR}
\resizebox{\linewidth}{!}{
\begin{tabular}{|c|c|c|c|c|c|c|c|}
\hline
Datasets         &NT & $T_{mmd}^{nk}$    & $T_{mmd}^{lin}$ & $T_{mmd}^{rbf}$ & $T_{m3d}^{nk}$ & $T_{m3d}^{lin}$ &$T_{m3d}^{rbf}$ \\ \hline \hline
$\mathbf{D}_1$ &2.833	&2.589	&2.612	&2.706	&1.476	&1.492	&\textbf{0.667}  \\ \hline
$\mathbf{D}_2$ &2.615	&1.899	&2.086	&2.650	&1.678	&1.669	&\textbf{0.990}  \\ \hline
$\mathbf{D}_3$ &2.546	&2.366	&2.331	&2.337	&1.515	&1.465	&\textbf{0.543}  \\ \hline
$\mathbf{D}_4$ &2.712	&2.357	&2.414	&2.505	&1.551	&1.480	&\textbf{0.534}  \\ \hline
$\mathbf{D}_5$ &2.536	&2.058	&2.310	&2.385	&1.557	&1.581	&\textbf{1.183}  \\ \hline
$\mathbf{D}_6$ &2.224	&1.838	&1.974	&2.105	&1.637	&1.637	&\textbf{0.731}  \\ \hline
$\mathbf{D}_7$ &1.843 &1.852	&1.639	&1.786	&1.556	&1.525	&\textbf{0.436}  \\ \hline
$\mathbf{D}_8$ &3.055	&2.674	&2.702	&2.685	&1.396	&1.402	&\textbf{0.566}  \\ \hline
$\mathbf{D}_9$ &2.224	&2.024	&2.020	&1.760	&1.725	&1.715	&\textbf{0.644}  \\ \hline
$\mathbf{D}_{10}$ &1.554 &1.701 &1.273 &1.371	&1.367	&1.330	&\textbf{0.686}  \\ \hline \hline
 Mean   &2.4141 &2.1359 &2.1360 &2.2290 &1.5459 &1.5295 &\textbf{0.6980}  \\ \hline
\end{tabular}}
\end{table}

In table \ref{ARCR}, we demonstrate the comparison results on $\mathbf{D}_i, i =1,...,10$, where NT is the no transfer baseline, $T_*^{nk}$ is the non-kernelized baseline, and $T_*^{lin}$ and $T_*^{rbf}$ are kernelized baselines using `linear' and `rbf' kernel, respectively.
Note that in this experiment, we set $N=5$ which is the true number of low-dimensional manifolds.
The discussion on the effect of $N$ on the transfer performance is presented in the next experiment.
From Table \ref{ARCR}, we can see that $T_{m3d}$ related baselines consistently outperform $T_{mmd}$ related ones in all the transfer tasks.
This is because $T_{m3d}$ minimizes the local distribution discrepancy in each manifold between two domains, and thus brings the two domains much closer than $T_{mmd}$ which only minimizes the global distribution discrepancy.
Regarding $T_{m3d}$ related baselines, we observe that $T_{m3d}^{rbf}$ yields the best results among all the baselines.
This is under expectation as `rbf' kernel has been shown effective in both conventional machine learning
and transfer learning, especially when no prior knowledge is available.
For the linear mapping cases $T_{m3d}^{nk}$ and $T_{m3d}^{lin}$, they achieve the comparable results.
In the following real-world experiments, we utilize a cross validation strategy to determine which linear strategy is used.
Specifically, we calculate the 5-fold training error using the source labelled data for each case, and select the one with the smaller source training error.
This is motivated by the fact that a good feature representation for domain adaptation should have small training error on the source domain

\begin{figure}[t]
\begin{center}
\includegraphics[scale=0.44]{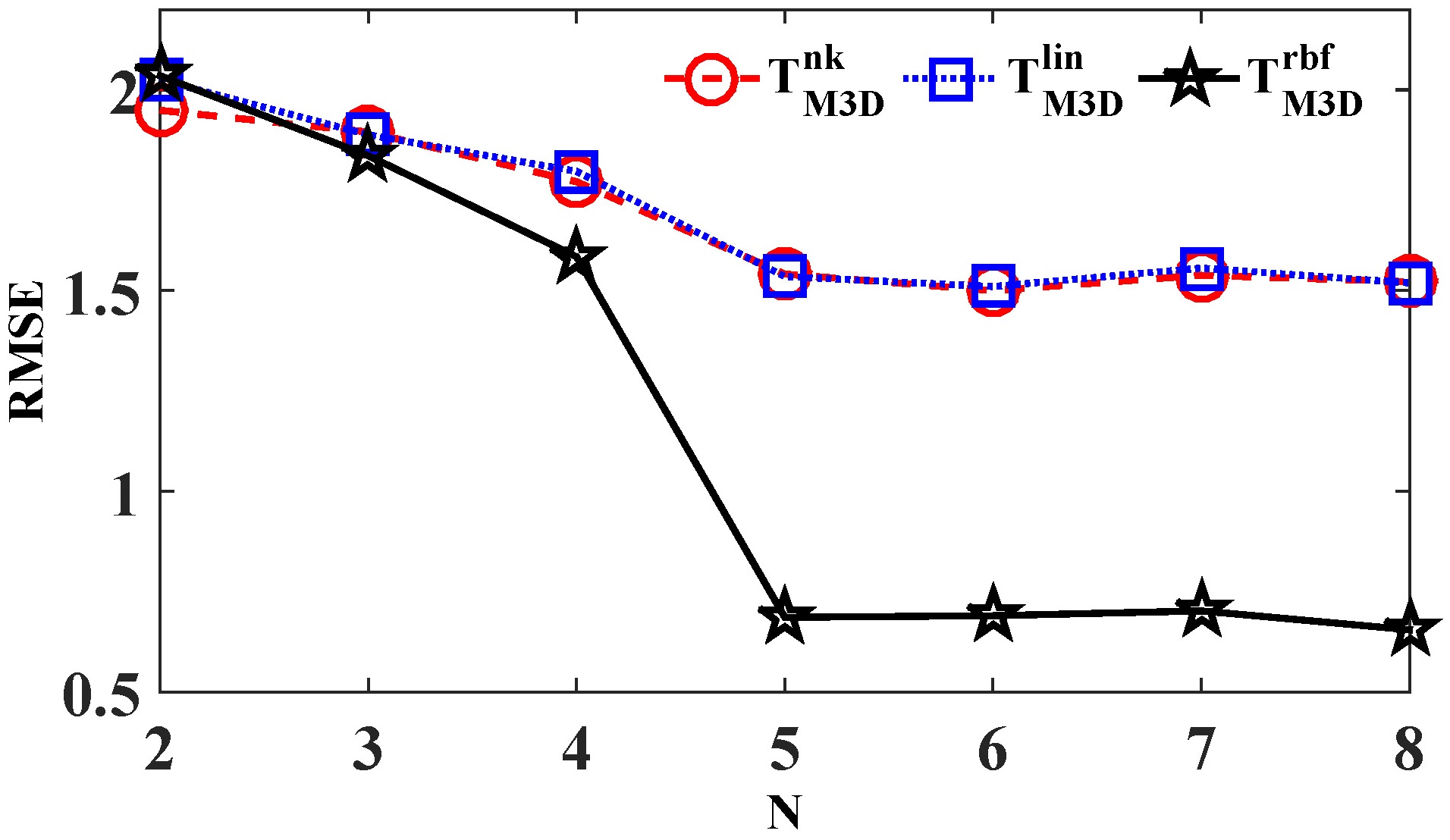}
\end{center}
\caption{RMSE with respect to $N$} \label{clusterNum}
\end{figure}

We then analyze how $N$, i.e., the number of manifolds set in the manifold clustering, affects the transfer performance.
We set $N = \{2,3,4,5,6,7,8\}$, and plot the average transfer results of $T_{m3d}^{nk}$, $T_{m3d}^{lin}$, and $T_{m3d}^{rbf}$ on $\mathcal{D}_1,...,\mathcal{D}_{10}$ in Figure \ref{clusterNum}.
It can be observed that the transfer performance generally becomes better as $N$ increases, and it achieves the optima when $N$ reaches the true number of manifolds, i.e., 5.
Afterwards, with the increase of $N$, the transfer performance gets stable.
This is reasonable as the local domain divergence are fully reduced if the manifolds are correctly clustered.

\subsection{Real-world Experiments}
We then compare TMDA with state-of-the-art transfer learning baselines, specifically in both the subspace learning and deep learning cases, on 4 real-world datasets.
Note that the first 2 datasets are benchmark datasets for the subspace-based transfer methods, and the latter 2 datasets are widely used for the evaluation of deep learning based transfer methods.
\cite{ben2007analysis}.

\noindent \textbf{COIL1-COIL2} \cite{long2013transfer} is an object recognition dataset consisting 20 objects.
It includes images from two domains \textbf{COIL1} and \textbf{COIL2}.
Each domain has 720 images, and each image is $32\times32$ pixels with 256 gray levels per pixel.
The images in two domains are taken in different directions, and thus are drawn from different distributions.
\textbf{COIL1} contains images taken in the directions of $[0^\circ,85^\circ]\cup[180^\circ,265^\circ]$.
\textbf{COIL2} contains images taken in the directions of $[90^\circ,175^\circ]\cup[270^\circ,355^\circ]$.
We construct two adaptation tasks: COIL$_{12}$ and COIL$_{21}$.

\noindent \textbf{20-Newsgroups} \cite{dai2007co} is a text dataset consisting of four top topics: computer (C), recording (R), science (S), and talk (T).
Each topic has four subtopics.
Following \cite{Wei2016DeepNF}, we use top topics as labels and form related domains using subtopics.
For instance, given two top topics C and R, top topic C is the positive label and top topic R is the negative label.
Two subtopics under C and R are selected to form one domain, and another two subtopics constitute another domain.
By pairing up the two domains, we have two adaptation tasks: CR-1 and CR-2.

\noindent \textbf{Office-31} \cite{long2015learning} is a benchmark dataset for visual domain adaptation. It contains 4,652 images and 31 categories from three distinct domains: Amazon (A), Webcam (W) and DSLR (D). By pairing up two domains, we construct 6 transfer tasks.

\noindent \textbf{ImageCLEF-DA} \cite{Long2017Deep} is built for ImageCLEF 2014 domain adaptation challenge. It contains 4 domains including Caltech-256 (C), ImageNet ILSVRC 2012 (I), Bing (B) and Pascal VOC 2012 (P). Following \cite{Long2017Deep}, we construct 6 transfer tasks.
\begin{table*}[!t]
\caption{Comparison results of the subspace learning case} \label{cr}
\resizebox{\linewidth}{!}{
\renewcommand\arraystretch{1.1}
\begin{tabular}{|c|c|c|c|c|c|c|c|c|c|c|c|c|}
\hline
Dataset & NN    & TCA   & SA    & GFK  & JDA  & TJM   & COR   & JGSA  & MEDA  & MMIT & TMDA$_{l}$ & TMDA$_{nl}$ \\ \hline \hline
COIL$_{12}$ & 80.97 & 84.72 & 57.92 & 78.19 & 84.31 & 78.19 & 88.33 & 86.39 & 65.56 & 89.86         & 92.25$\pm$0.86      & \textbf{92.25$\pm$0.05}      \\ \hline
COIL$_{21}$ & 81.53 & 84.03 & 44.58 & 79.44 & 87.08 & 78.75 & 85.00 & 86.81 & 66.67 & 90.56         & 92.53$\pm$0.95      & \textbf{92.78$\pm$0.01}     \\ \hline \hline
CR-1        & 56.90 & 57.84 & 59.04 & 59.64 & 63.41 & 57.93 & 55.27 & 59.73 & 50.73 & 65.81         & 73.01$\pm$0.01       & \textbf{77.10$\pm$0.12}      \\ \hline
CR-2        & 51.23 & 58.86 & 58.10 & 58.27 & 66.84 & 62.00 & 57.68 & 60.05 & 50.81 & 78.54          & 88.13$\pm$0.00      & \textbf{88.74$\pm$0.12}    \\ \hline
CS-1        & 60.31 & 62.36 & 62.70 & 62.19 & 67.15 & 61.76 & 56.80 & 64.59 & 50.81 & 68.95          & 76.24$\pm$0.00      & \textbf{76.97$\pm$0.10}       \\ \hline
CS-2        & 54.51 & 61.41 & 63.88 & 61.24 & 64.65 & 62.35 & 55.20 & 62.44 & 50.51 & 62.10          & 70.27$\pm$0.00      & \textbf{73.53$\pm$0.39}      \\ \hline
CT-1        & 59.62 & 74.26 & 74.45 & 74.16 & 75.12 & 77.99 & 60.10 & 74.83 & 55.60 & 80.67         & 84.54$\pm$0.02     & \textbf{88.06$\pm$0.07}      \\ \hline
CT-2        & 58.78 & 71.50 & 73.50 & 71.29 & 73.40 & 75.29 & 64.35 & 74.03 & 60.46 & 86.86         & 90.68$\pm$0.02      & \textbf{91.50$\pm$0.04}     \\ \hline
RS-1        & 59.29 & 63.08 & 64.42 & 64.17 & 68.04 & 61.90 & 59.63 & 65.26 & 50.13 & 70.40          & \textbf{83.47$\pm$0.44}      & 76.18$\pm$0.19      \\ \hline
RS-2        & 62.31 & 63.91 & 62.56 & 66.27 & 66.78 & 64.33 & 62.39 & 64.76 & 50.08 & 76.64         & \textbf{85.97$\pm$0.30}     & 81.45$\pm$0.06    \\ \hline
RT-1        & 51.89 & 59.28 & 63.07 & 62.88 & 66.48 & 65.72 & 52.84 & 64.96 & 56.16 & 72.35         & 74.34$\pm$0.00     & \textbf{76.16$\pm$0.04}    \\ \hline
RT-2        & 53.49 & 54.52 & 62.42 & 59.75 & 63.45 & 66.63 & 58.52 & 64.07 & 61.40 & 74.64         & 79.16$\pm$0.00      & \textbf{79.67$\pm$0.03}      \\ \hline
ST-1        & 54.78 & 63.39 & 64.52 & 64.24 & 64.33 & 62.82 & 57.05 & 64.52 & 56.39 & 73.70         & 76.65$\pm$0.01      & \textbf{78.85$\pm$0.16}       \\ \hline
ST-2        & 57.39 & 62.36 & 62.77 & 63.19 & 62.56 & 64.94 & 58.63 & 65.56 & 61.12 & 80.46          & 84.28$\pm$0.00      & \textbf{85.01$\pm$0.14}      \\ \hline \hline
Average     & 60.21 & 65.82 & 62.43 & 66.07 & 69.54 & 67.19 & 62.27 & 68.43 & 56.17 & 76.54         & 82.25$\pm$0.19      & \textbf{82.73$\pm$0.11}    \\ \hline
\end{tabular}}
\end{table*}

\subsubsection{Subspace Learning Comparisons}
Regarding the subspace learning case, we compare various subspace-based baselines including TCA \cite{pan2011domain}, GFK \cite{gong2012geodesic}, SA \cite{fernando2013unsupervised}, JDA\cite{long2013transfer}, TJM \cite{long2014transfer}, COR \cite{sun2016return}, JGSA \cite{zhang2017joint}, MEDA \cite{wang2018visual}, and MMIT \cite{wei2019knowledge} on the first two datasets.
For the methods using nonlinear kernels, we use the default `rbf' kernel.
As suggested by \cite{gong2012geodesic}, Nearest Neighbor (NN) classifier is used as the base classifier.
We utilize Maximum Likelihood Estimator \cite{levina2004maximum} to determine the dimensionality of new representations in all the methods for fair comparison.
For the hyper-parameters in each baseline, we use the default values specified by the authors.
Regarding our TMDA, we follow hyper-parameter instructions stated in the technical section, and empirically set $\alpha = 0.01$ and $\beta = 100$.
We set the maximum number of iterations as 50 in the optimization.
As the true number of manifolds for each transfer task is unknown, we set $N$ as the number of classes in the source domain, for both datasets.
To obtain statistical results, we run 5 times and take the mean with variance as the final result.

Table \ref{cr} shows the comparison results of the subspace learning case.
TMDA$_{l}$ and TMDA$_{nl}$ represent TMDA with linear mapping and nonlinear mapping, respectively.
Specifically, we use the cross validation strategy on the source domain to select the linear mapping form as stated in Section 7.1.
For nonlinear mapping, we use the kernelized TMDA with `rbf' kernel.
The best result in each transfer task is highlighted using bold.
As can be seen in table \ref{cr}, all the best results are located in the TMDA related methods.
Specifically, TMDA$_{nl}$ achieves 12 best transfer performance out of 14 transfer tasks.
Regarding the average performance on all the transfer tasks, TMDA$_{nl}$ beats all the baselines, and is the clear winner.
Note that TMDA$_{nl}$ yields 16.91\% improvement compared with TCA, which demonstrates that the alignment of local distribution discrepancy of subdomains is much more efficient than the alignment of the global distribution discrepancy.
Moreover, TMDA$_{nl}$ achieves 13.19\% average improvement compared with JDA.
As JDA is using CMMD, it demonstrates us the superiority of using manifold to class in the definition of subdomains.
All the results indicate that TMDA is an effective transfer learning method in the subspace learning case.
\begin{table}[t]
\centering
\caption{Comparison results of office31} \label{off31}
\begin{tabular}{|c|c|c|c|c|c|c|c|}
\hline
Tasks    & W-A  & A-W  & D-A  & A-D  & W-D  & D-W  & Average  \\ \hline \hline
RESNET50 & 60.7 & 68.4 & 62.5 & 68.9 & 99.3 & 96.7 & 76.1 \\ \hline
DAN      & 62.8 & 80.5 & 63.6 & 78.6 & 99.6 & 97.1 & 80.4 \\ \hline
DANN     & \textbf{67.4} & 82.0 & \textbf{68.2} & 79.7 & 99.1 & 96.9 & 82.2 \\ \hline
RTN      & 64.8 & 84.5 & 66.2 & 77.5 & 99.4 & 96.8 & 81.6 \\ \hline
CAN      & 63.4 & 81.5 & 65.9 & \textbf{85.5} & 99.7 & \textbf{98.2} & 82.4 \\ \hline
TMDA     & 64.8 & \textbf{86.2} & 65.7 & 83.3 & \textbf{99.8} & 97.1 & \textbf{82.8} \\ \hline
\end{tabular}
\end{table}

\subsubsection{Deep Learning Comparisons}
For the deep learning case, we compare the state-of-the-art deep transfer methods including DAN \cite{long2015learning}, DANN \cite{ganin2014unsupervised}, RTN \cite{long2016unsupervised}, CAN \cite{zhang2018collaborative} on the last two datasets.
We use PyTorch DAN \cite{long2015learning} as the base framework for the implementation of our TMDA.
The pre-trained ResNet50 model is used as the CNN feature extractor.
We use stochastic gradient descent (SGD) for optimization.
The learning rate decreases gradually after each iteration from 0.01, and we adopt the same learning
rate decrease strategy as in \cite{ganin2014unsupervised}.
We set the epochs, batch size, momentum, and weight decay as 200, 32, 0.9, and $5\times10^{-4}$, respectively.
Regarding the hyper-parameters $\lambda_1$ and $\lambda_2$, we gradually update them from 0 to 1 by a progressive schedule \cite{Long2017Deep}, $\lambda_i = 2 / (1 + exp(-5 * (200-p) / 200)) -1$ where $p$ is the current epoch number, to reduce the parameter sensitivity and ease the model selections.
\begin{table}[t]
\centering
\caption{Comparison results of ImageCLEF} \label{ImageCLEF}
\begin{tabular}{|c|c|c|c|c|c|c|c|}
\hline
Tasks    & C-I  & I-C  & I-P  & P-I  & C-P  & P-C  & Average  \\ \hline \hline
RESNET50 & 78.0 & 91.5 & 74.8 & 83.9 & 65.5 & 91.2 & 80.7 \\ \hline
DAN      & 86.3 & 92.8 & 74.5 & 82.2 & 69.2 & 89.8 & 82.5 \\ \hline
DANN     & 87.0 & \textbf{96.2} & 75.0 & 86.0 & 74.3 & 91.5 & 85.0 \\ \hline
RTN      & 86.9 & 95.3 & 75.6 & 86.8 & 72.7 & 92.2 & 84.9 \\ \hline
CAN      & \textbf{89.5} & 94.2 & 78.2 & 87.5 & 75.8 & 89.2 & 85.7 \\ \hline
TMDA     & 88.7 & 94.5 & \textbf{78.7} & \textbf{92.3} & \textbf{75.8} & \textbf{91.5} & \textbf{86.9} \\ \hline
\end{tabular}
\end{table}

Tables \ref{off31} and \ref{ImageCLEF} show the comparison results on office31 dataset and ImageCLEF dataset for the deep learning case, respectively.
Compared with the baselines, our proposed TMDA achieves competitive results regarding the average performance on both datasets.
We also note that our proposed TMDA may not be optimal compared with the most recent deep transfer learning methods.
However, we want to highlight the scalability of our TMDA framework, which can be built upon the existing deep transfer methods.
In this work, our aim is to show the effectiveness of our TMDA idea, i.e., sudomain alignment, on the deep learning case, and thus we instantiate the TMDA framework on the very benchmark deep transfer methods DAN.
It can be seen that it yields 2.4\% and 4.4\% average improvements than DAN from tables \ref{off31} and \ref{ImageCLEF}.
Considering that DAN uses the conventional MMD to align the global distribution discrepancy while TMDA exploits M3D to align the local distribution discrepancy in each manifold, the improvements indicate the effectiveness of the subdomain alignment using manifolds.
Moreover, we emphasize that TMDA can be readily applied to other deep transfer learning frameworks.
The principle is, for MMD-based methods, to take advantage of M3D instead of the conventional MMD.
Regarding the adversarial-based methods, we can instantiate our TMDA by constructing the domain discriminator in each manifold.
We leave these potential research topics in future works.

\subsection{Property Study}
In this section, we further analyze the properties of TMDA including the manifold number analysis, ablation study, and sensitivity analysis. We mainly focus on the subspace learning case, more specifically, nonlinear mapping with `rbf' kernel.

\subsubsection{Manifolds Number Analysis}
In this section, we analyze how the number of manifolds set in TMDA affects the final transfer performance.

\begin{figure}[!t]
\centering
{\includegraphics[scale = 0.5]{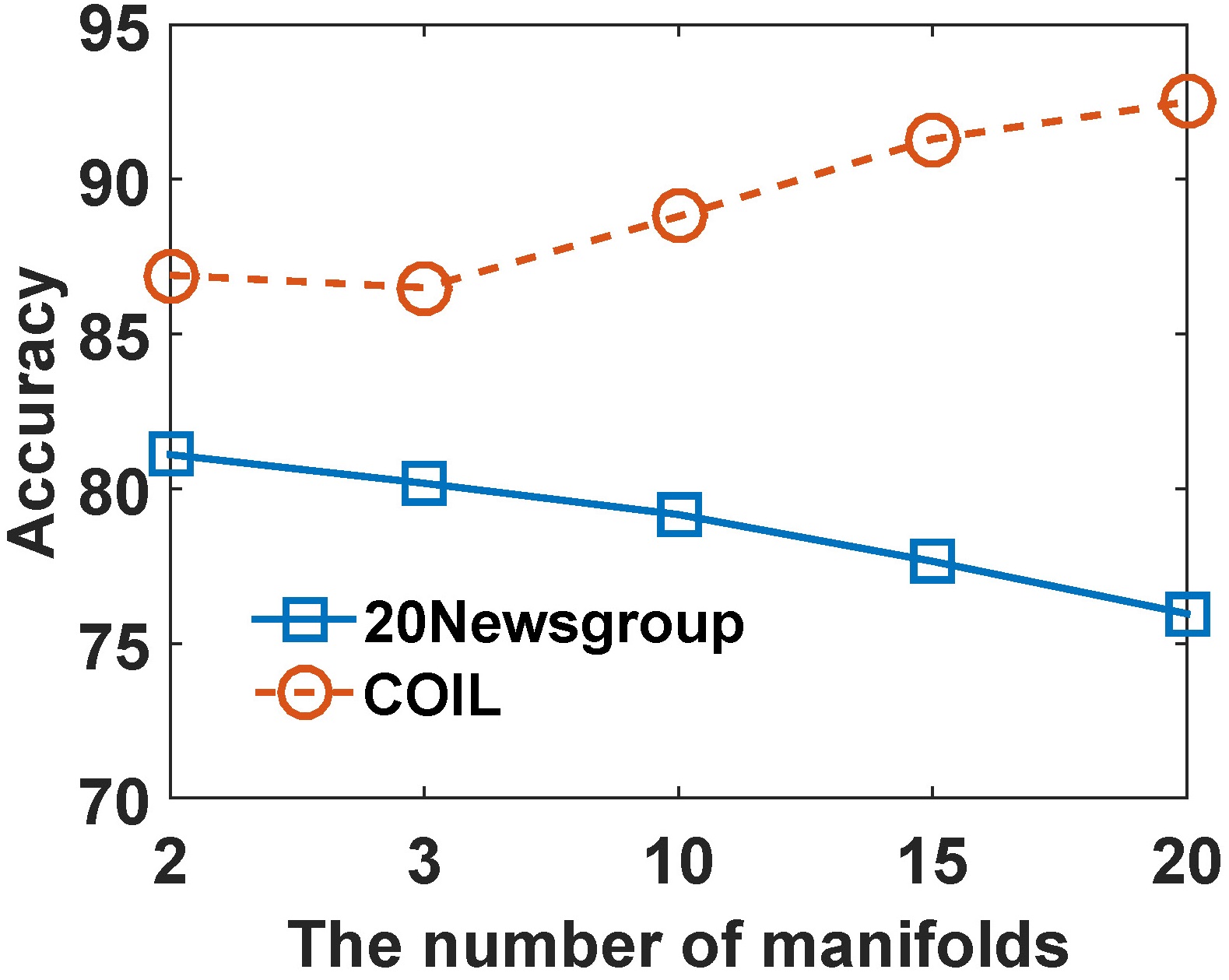} }
 \caption{Manifold number analysis of TMDA}  \label{manifods}
\end{figure}

Figure \ref{manifods} shows the average TMDA results of all the tasks for 20-Newsgroups and COIL datasets with the number of manifolds, $N$, varying in a range $[2,5,10,15,20]$.
It can be observed that, with the increase of $N$, the performance improves on COIL dataset, but slightly decays on 20-Newsgroups dataset.
This is because the two domains in 20-Newsgroups dataset consist of documents from two topics, and the true number of manifolds may be exactly 2.
However, for COIL dataset, it contains multiple objects with different backgrounds, and thus may consist of much more ($\geq 2$) low-dimensional manifolds.
Since we usually do not have any prior knowledge on the true number of manifolds given a specific dataset, we propose to set $N$ to be equal to the number of classes for TMDA.
Moreover, we also observe that when $N=2$, which is the smallest value, TMDA still achieves very promising results, i.e., 81.10\% for 20-newsgroup dataset and 86.93\% for COIL dataset.
This shows the effectiveness of using this compromised $N$, and indicates another alternative choice of setting $N$.

\subsubsection{Ablation Studies}
\begin{table}[t]
\centering
\caption{Ablation study of TMDA} \label{Ablation}
\resizebox{\linewidth}{!}{
\begin{tabular}{|c|c|c|c|c|}
\hline
Dataset & NN    & TMDA\_$v1$    & TMDA\_$v2$    & TMDA  \\ \hline \hline
COIL$_{12}$  & 80.97 & 88.34 & 88.72 & 92.25 \\ \hline
COIL$_{21}$  & 81.53 & 88.92 & 91.94 & 92.78 \\ \hline \hline
CR-1     & 56.90 & 70.30 & 75.99 & 77.10 \\ \hline
CR-2     & 51.23 & 79.13 & 88.04 & 88.74 \\ \hline
CS-1     & 60.31 & 70.86 & 77.79 & 76.97 \\ \hline
CS-2     & 54.51 & 67.34 & 72.59 & 73.53 \\ \hline
CT-1     & 59.62 & 82.66 & 86.62 & 88.06 \\ \hline
CT-2     & 58.78 & 88.52 & 91.50 & 91.50 \\ \hline
RS-1     & 59.29 & 71.49 & 76.00 & 76.18 \\ \hline
RS-2     & 62.31 & 77.27 & 81.26 & 81.45 \\ \hline
RT-1     & 51.89 & 70.66 & 75.97 & 76.16 \\ \hline
RT-2     & 53.49 & 71.25 & 79.26 & 79.67 \\ \hline
ST-1     & 54.78 & 77.45 & 78.26 & 78.85 \\ \hline
ST-2     & 57.39 & 81.12 & 83.64 & 85.01 \\ \hline \hline
Average & 60.21 & 77.52 & 81.97 & 82.73 \\ \hline
\end{tabular}}
\end{table}
In this section, we conduct the ablation studies by comparing TMDA with two variants of TMDA.
The first variant, TMDA\_$v1$, follows the optimization objective (\ref{Eq4}) but replaces M3D with the conventional MMD.
In another word, instead of aligning the domain divergence in each subdomain, TMDA\_$v1$ directly aligns the global domain divergence.
The second variant, TMDA\_$v2$, decouples the discovery of low dimensional manifolds and the adaptation process.
More specifically, we separate the optimization problem (\ref{Eq4}) into the following two sub-problems:
\emph{sub-problem (1)}:
\begin{equation} \nonumber
    \begin{array}{*{20}{c}}
   {\mathop {\min }\limits_{\{ {A_{ij}}\} } {\rm{   }}\sum\limits_i {||{{\bf{x}}_i} - \sum\limits_j {{A_{ij}}{{\bf{x}}_j}} |{|^2} + \mu |{A_{ij}}|} }  \\
   {s.t.{\rm{ \ \ \  }}{\rm{  diag(}}{\bf{A}}{\rm{) = }}{\bf{0}},}  \\
   \end{array}
\end{equation}

\noindent and \emph{sub-problem (2)}:
\begin{eqnarray} \nonumber
\begin{aligned}
   {\mathop {\min }\limits_{\bf{W}}} {\rm{\ \ \ \ }} &  \alpha ||\mathbf{W}^{\mathsf{T}}\mathbf{K}-\mathbf{W}^{\mathsf{T}}\mathbf{K}\mathbf{A}||_F^2 + \frac{\beta}{2} tr(\mathbf{W}^{\mathsf{T}}{\bf{K}}({\sum\limits_{m=1}^N\mathbf{M}^m}){{\bf{K}}^{\mathsf{T}}}\mathbf{W}) \\ & + tr(\mathbf{W}^{\mathsf{T}}\mathbf{W}) \\
    s.t.{\rm{\ \ \ \ }} &{{\bf{W}}^{\mathsf{T}}}{\bf{X}}{{\bf{X}}^{\mathsf{T}}}{\bf{W}} = \mathbf{I}.
\end{aligned}
\end{eqnarray}
The \emph{sub-problem (1)} pre-learns the sparsest $\mathbf{A}$, and then \emph{sub-problem (2)} uses the learned $\mathbf{A}$ to align subdomain divergence for transfer.
Note that TMDA\_$v2$ utilizes the sparest $\mathbf{A}$ that may be not with the best transfer capacity.
We test TMDA with these two variants on both COIL and 20-Newsgroups datasets.
The results of no-transfer baseline are also shown as references.

Table \ref{Ablation} shows the comparison results.
Overall, TMDA outperforms TMDA\_$v1$ and TMDA\_$v2$.
Compared TMDA with TMDA\_$v1$, it can be concluded that the subdomain alignment brings more benefits to transfer than the global domain alignment.
This verifies the effectiveness of exploring the subdomains in transfer.
Compared TMDA with TMDA\_$v2$, the improvements indicate that the joint learning strategy can discover the low-dimensional manifolds with a better transfer capacity than the decoupled one.
This supports the necessity of using the joint learning in TMDA to adaptation.
All the comparison results show the effectiveness of TMDA to transfer learning problems.
\begin{figure}[t]
\begin{center}
\includegraphics[scale=0.45]{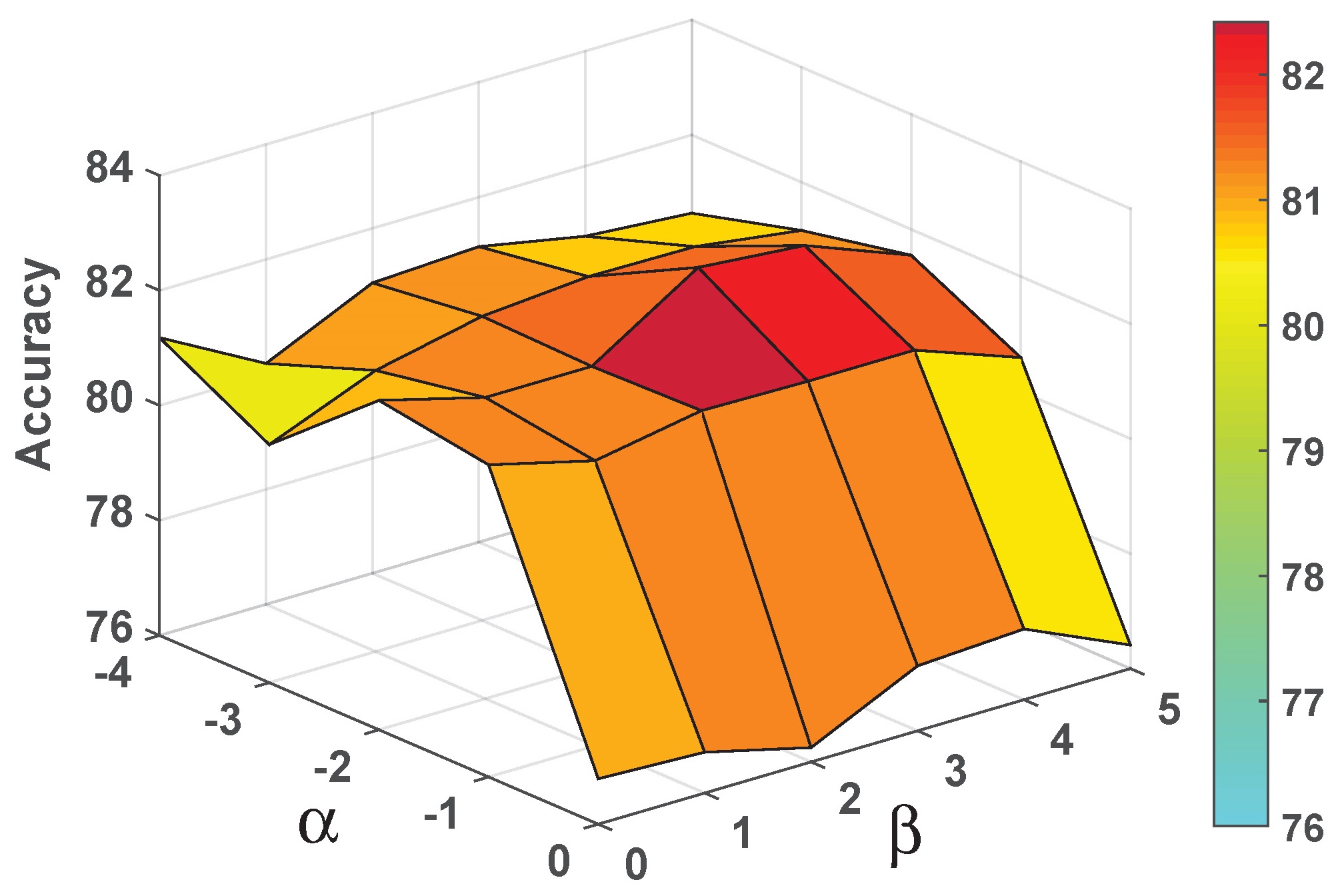}
\end{center}
\caption{Sensitivity analysis} \label{Sensitivity}
\end{figure}

\subsubsection{Sensitivity Analysis}
We conduct the sensitivity analysis on the two hyper-parameters $\alpha$ and $\beta$.
Figure \ref{Sensitivity} shows the average results of all the subspace learning tasks with $\alpha$ and $\beta$ varying in different ranges ($\alpha$ =$log_{10}^{x}$ and $\beta$ =$log_{10}^{y}$).
We observe that the best results appear from $\alpha = 0.01$ and $\beta = 100$.
In the experiments of the subspace learning case, we empirically set $\alpha = 0.01$ and $\beta = 100$ for all the tasks.


\section{Conclusions and Future Works}
In this paper, we consider the local subdomain divergence across domains in transfer.
Specifically, we define subdomains using manifolds, and propose to align the local distribution discrepancy in each manifold.
A new metric, Manifold Maximum Mean Discrepancy (M3D), and a general transfer framework, Transfer with Manifolds Discrepancy Alignment (TMDA), are developed.
We instantiate TMDA in the subspace learning case considering both the linear and nonlinear mappings.
We also instantiate TMDA in deep learning based on DAN framework.
We compare our TMDA with the state-of-the-art subspace and deep learning transfer baselines.
Extensive comparison results show that TMDA is a very promising method for transfer learning.

We emphasize that the idea of TMDA is general in the sense that it can be instantiated using different techniques.
We propose one instantiation based on subspace clustering, MMD, and Frobenious norm in this work, and leave the other possibilities for future exploration.
Moreover, in this work, the instantiation of TMDA for deep learning case is based on the very benchmark baseline, DAN, as our aim is to demonstrate its effectiveness in deep learning.
However, we highlight that the TMDA framework is scalable in the sense that it can be built upon some other deep learning transfer baselines.
We provide instructions for the application of TMDA to both MMD based and adversarial loss based deep transfer methods.
Considering the rapid development of deep transfer learning methods, we will study much more advanced instantiation of TMDA for deep learning in future works.

Finnaly, the idea of subdomain alignment using manifolds is applicable for many problem settings, e.g., disentangled domain-invariant representation learning where each disentangled latent factor can be taken as one low-dimensional manifold, and multi-task reinforcement learning where multiple tasks are assumed to decompose into several subtasks corresponding to multiple manifolds, and the knowledge transfer is enforced to be happened in each subtask.
We leave these potential research directions in the future studies.

\section*{Acknowledgments}
The majority of this work was done when the first author was studying at Nanyang Technological University (NTU).
This work is partially supported by an Academic Research Grant No. MOE2016-T2-2-068 from the Ministry of Education, Singapore.

\bibliographystyle{IEEEtran}
\bibliography{TMDA}

\end{document}